# Relative Loss Bounds for On-line Density Estimation with the Exponential Family of Distributions

**Katy S. Azoury**
College of Business
San Francisco State University
San Francisco, CA 94132
U.S.A
kazoury@sfsu.edu

**M. K. Warmuth**
Computer Science Department
University of California, Santa Cruz
Santa Cruz, CA 95064
U.S.A.
manfred@cse.ucsc.edu

## Abstract

We consider on-line density estimation with a parameterized density from the exponential family. The on-line algorithm receives one example at a time and maintains a parameter that is essentially an average of the past examples. After receiving an example the algorithm incurs a loss which is the negative log-likelihood of the example w.r.t. the past parameter of the algorithm. An off-line algorithm can choose the best parameter based on all the examples. We prove bounds on the additional total loss of the on-line algorithm over the total loss of the off-line algorithm. These relative loss bounds hold for an arbitrary sequence of examples. The goal is to design algorithms with the best possible relative loss bounds. We use a certain divergence to derive and analyze the algorithms. This divergence is a relative entropy between two exponential distributions.

## 1 Introduction

Consider density estimation with a Gaussian of fixed variance. The *off-line* algorithm is given a batch of examples $\{x_1, x_2, \ldots, x_T\}$ all at once and it chooses its mean parameter based on all $T$ examples. The choice of the mean which maximizes the likelihood of the examples is clearly the average of the examples. The *on-line* algorithm sees the examples one at a time. At trial $t$ it has to choose or predict with a mean $\mu_t$ based on the past $t-1$ examples. It then receives the example $x_t$ and incurrs a loss $(\mu_t - x_t)^2/2$. Finally, at the end of trial $t$, the on-line algorithm updates its parameter (hypothesis) from $\mu_t$ to $\mu_{t+1}$.

The performance of the on-line algorithm is measured by the total loss on the whole sequence of examples. The goal is to design on-line learning algorithms with good bounds on their total loss. Even though the motivation of the parameter updates may have a probabilistic interpretation, the bounds we prove hold for an arbitrary (or worst-case) sequence of examples. Clearly, there cannot be meaningful bounds on the total loss of an algorithm that stand alone as well as hold for an arbitrary sequences of examples. However, in on-line learning a certain type of "relative" loss bound is desirable and has been used successfully. The term "relative" means that there is a comparison to the best off-line model from a comparison class of models. The *relative loss bounds* quantify the additional total loss of the on-line algorithm over the total loss of the best off-line model. Since the on-line learner does not see the sequence of examples in advance, the relative loss bounds estimate the price of hiding the future examples from the learner.

The simplest on-line algorithm in the Gaussian example predicts at trial $t$ as the off-line algorithm would on the batch $\{x_1, x_2, \ldots, x_{t-1}\}$. That is, $\mu_t$ is the average of the past $t-1$ examples. We call this the *incremental off-line algorithm*. The purpose of this paper is to prove relative loss bounds for the incremental off-line algorithm and a variant of this algorithm that averages by $t$ instead of $t-1$ when computing $\mu_t$. We call the latter algorithm the *forward algorithm*

The bounds are developed for density estimation with a parameterized distribution from the exponential family. The Gaussian density is just a special case. In the general case the mean parameters remain averages or modified averages of the past instances.

The forward algorithm was inspired by the work of Vovk [Vov97] on linear regression which can also be interpreted as a density estimation problem. For linear regression the incremental off-line algorithm becomes the on-line linear least squares algorithm or ridge regression. Dividing by $t-1$ for computing $\mu_t$ corresponds to multiplying with the inverse covariance matrix of the first $t-1$ examples. The forward algorithm for linear regression multiplies with the inverse covari-ance matrix of the first $t$ examples instead.

## Overview of previous work

This work is rooted in the study of relative loss bounds for on-line learning algorithms of the computational learning theory community. Even though these bounds may underestimate the performance on natural data we have been using them as a powerful yardstick for analyzing and comparing algorithms. This line of research was initiated by Littlestone with the discovery of the Winnow algorithm [Lit88]. He also pioneered a style of amortized analysis for proving relative loss bounds which use certain divergence functions as potential functions. Winnow is designed for disjunctions as the comparison class and the total number of mistakes is used as the loss. The next wave of on-line algorithms was designed for a finite set of experts as a comparison class and a wide range of loss functions [LW94, Vov90, CBFH+97, HKW98]. After that, algorithms were developed for the on-line linear least squares regression, i.e. when the comparison class consists of linear neurons (linear combination of experts) [LLW95, CBLW96, KW97]. This work has been generalized to the case when sigmoided linear neurons are the comparison class [HKW95, KW98]. General frameworks of on-line learning algorithms were developed in [GLS97, KW97, KW98]. We follow the philosophy of Kivinen and Warmuth [KW97] of starting with a divergence function. From the divergence function we derive the on-line update and we then use the same divergence as a potential in the amortized analysis. A similar method was developed in [GLS97] for the case when the comparison class consists of linear threshold functions. They start with an update and construct the appropriate divergence that is used in the analysis.

Recently we have been discovering that the divergences that were used in on-line learning have been employed extensively in convex optimization and are called Bregman-distances [Bre67, CL81, Csi91, JB90]. We use the term "divergence" instead of "distance" since the triangular inequality does not hold for such divergences. Bregman divergences can also be seen as relative entropies between distributions from the exponential family of distributions [Ama85]. In particular a notion of duality was explored by Amari that is inherent in these divergences. In a later section we bring out the connection to the exponential family of distributions and also discuss briefly how the Bregman divergences are used in convex optimization.

We use the divergence to prove a different type of relative loss bounds that grow logarithmically with the number of trials $T$. These types of bounds have been investigated in a Bayesian context by [Fre96, XB97, Yam98] where a generalized posterior is maintained on all parameters. Such bounds have also been proven for on-line linear regression [Fos91, Vov97]. An important insight we gained from this research is that $\log(T)$-style bounds can be obtained by using variable learning rates. This is a key feature of this work and distinguishes it from most previous work on relative loss bounds which typically uses the same learning rate in each trial. In this paper the learning rate applied in trial $t$ is proportional to $1/t$.

Our updates are simple and operate directly on the sufficient statistic of the parametric exponential family. Our methodology applies to any member of the exponential family of distributions. In summary, a key goal of this paper is to stress the importance of the divergences as a tool for deriving and analyzing on-line learning algorithms and to bring out the advantage of variable learning rates.

## Outline

In the next section we present an overview of important properties of the exponential family and introduce the notion of divergences and their duality properties. In Section 3 we treat density estimation for the exponential family. We give a number of examples of how our methods can be applied to particular members of the exponential family. We then outline in Section 4 how the relative loss bounds for linear regression that where proven by Vovk [Vov97] can be obtained with our methods. In Section 5 we briefly discuss Vovk's alternate method for proving relative loss bounds that uses integration over a generalized posterior. Our techniques are more concise. Finally we conclude (Section 6) with a discussion of future work.

## 2 Exponential Families of Distributions

In this section we review some important features of the exponential family of distributions that will be used in this paper. The features of interest that are used throughout this paper include measures of divergence between two members of the family and an intrinsic duality relationship. See [BN78, Ama85] for a more comprehensive treatment.

A multivariate parametric family $\mathcal{F}_G$ of distributions is said to be an exponential family, when its members have a density function of the form

$$P_G(x|\theta) = \exp(\theta \cdot x - G(\theta))P_0(x) \quad (2.1)$$

where $\theta$ and $x$ are vectors in $R^d$, $\theta \cdot x$ denotes the dot product, and $P_0(x)$ represents any factor of the density which does not depend on $\theta$.

The $d$-dimensional parameter $\theta$ is usually called the *natural* (or *canonical*) parameter. Many common parametric distributions are members of this family, including the Gaussian. The function $G(\theta)$ is a normalization factor defined by

$$G(\theta) = \ln \int (\exp(\theta \cdot x)) P_0(x) dx$$

The space $\Theta \subseteq R^d$ for which the integral above is finite is called the *natural* parameter space. The exponential family is called *regular* if $\Theta$ is an open subset of $R^d$. It is well known that $\Theta$ is a convex set, and that $G(\theta)$ is a strictly convex function on $\Theta$. Moreover, $G(\theta)$, is called the cumulant function of (2.1), and it plays a fundamental role in characterizing members of this family of distributions.

We use $g(\theta)$ to denote the gradient vector $\nabla_\theta G(\theta)$, and $G_{ij}(\theta)$ to denote the matrix of second partials of the cumulant function $G(\theta)$. Let

$$\lambda_G(\theta; x) = \ln P_G(x|\theta)$$

represent the log-likelihood which is viewed as a function of $\theta$. Under some standard regularity conditions, log-likelihood functions, in general, satisfy some well known moment identities [MN89]. Applying these identities to the exponential family reveals the special role played by the cumulant function $G(\theta)$.

Let $E_\theta(\cdot)$ be the expectation with respect to the distribution $P_G(x|\theta)$. The first moment identity of log-likelihood functions is

$$E_\theta(\nabla_\theta \lambda_G(\theta; x)) = 0 \quad (2.2)$$

For the exponential family, this identity implies that

$$E_\theta(x) = \nabla_\theta G(\theta) = g(\theta)$$

thus, showing that the gradient $g(\theta)$ is equal to the mean of $x$. We let $g(\theta) = \mu$, and call $\mu$ the expectation parameter.

Since the cumulant function $G$ is convex, the map $g(\theta) = \mu$ has an inverse. We denote the image of $\Theta$ under the map $g(.)$ by $\mathbf{M}$ and the inverse map from $\mathbf{M}$ to $\Theta$ by $g^{-1}(\mu) = \theta$. The set $\mathbf{M}$ is called the expectation space, which may not necessarily be a convex set.

The second moment identity for log-likelihood functions is

$$E_\theta \left( \nabla_\theta \lambda_G(\theta; x)(\nabla_\theta \lambda_G(\theta; x))' \right)$$
$$= E_\theta \left( (x - \mu)(x - \mu)' \right)$$
$$= -E_\theta \left( \frac{\partial^2}{\partial \theta^2} \lambda_G(\theta; x) \right).$$

Here $'$ denotes the transpose operation.

For the exponential family we have that

$$-\frac{\partial^2}{\partial \theta^2} \lambda_G(\theta; x) = \frac{\partial^2}{\partial \theta^2} G(\theta) = G_{ij}(\theta).$$

Thus the Hessian of $G(\theta)$ (the matrix $G_{ij}(\theta)$) is the variance-covariance matrix $\text{Var}(x|\theta)$ for $x$. This matrix is also called the *Fisher Information Matrix*.

### 2.1 Duality Between the Natural Parameters and the Expectation Parameters

Sometimes, it is more convenient to parametrize a distribution in the exponential family by using its expectation parameter $\mu$ instead of its natural parameter $\theta$. This pair of parametrizations have a dual relationship. We provide the aspects of the duality that are relevant to this paper. Define a second function on the range $\mathbf{M}$ of $g$ as follows:

$$F(\mu) := \theta \cdot g(\theta) - G(\theta) \quad (2.3)$$

Let $f(\mu) = \nabla_\mu F(\mu)$ denote the gradient of $F(\mu)$.

Note that by taking the gradient of $F(\mu)$ in (2.3) with respect to $\mu$ and treating $\theta$ as a function of $\mu$, we get

$$\nabla_\mu F(\mu) = f(\mu) = \theta + \frac{\partial \theta}{\partial \mu} \mu - \frac{\partial \theta}{\partial \mu} g(\theta). \quad (2.4)$$

The last two terms cancel, and $f(\mu)$ is exactly the inverse map $g^{-1}(\mu)$. Since $g(\theta)$ has a positive definite Jacobian for all $\theta \in \Theta$, $g^{-1}(\mu)$ has a positive definite Jacobian for all $\mu \in \mathbf{M}$. Thus the second function $F(\mu)$ is strictly convex as well and the functions $G(\theta)$ and $F(\mu)$ form a pair of *dual* convex functions.

The two parametrizations $\theta$ and $\mu$ are related by the Legendre Transformations

$$\nabla_\theta G(\theta) = g(\theta) = \mu,$$
$$\text{and } \nabla_\mu F(\mu) = f(\mu) = \theta \quad (2.5)$$

Let $F_{ij}(\mu)$ denote the Hessian of $F(\mu)$. It follows from (2.5) that $F_{ij}(\mu)$ is equal to the inverse of the *Fisher Information Matrix*, since an immediate consequence of (2.5) is

$$\frac{\partial \mu}{\partial \theta} = G_{ij}(\theta)$$
$$\frac{\partial \theta}{\partial \mu} = F_{ij}(\mu) = G_{ij}^{-1}(\theta).$$

Now consider the function $V(\mu) = G_{ij}(f(\mu))$, which is the *Fisher Information Matrix* expressed in terms of the expectation parameter. This function is defined on the expectation space $\mathbf{M}$, takes values in the space of symmetric $d \times d$ matrices, and is called the *variance function*. The matrix $V(\mu)$ is positive definite for all $\mu \in \mathbf{M}$, and $V(\mu) = F_{ij}^{-1}(\mu)$. Thus in the context of exponential families the functions $F$ and $G$ are not arbitrary strictly convex functions but must have positive definite symmetric Hessians. However all the dual relationships hold in general for the case when $G(\theta)$ is an arbitrary differentiable convex function. In that case the dual function $F(\mu)$ is also called the convex conjugate [Roc70].

## 2.2 Divergence Between two Exponential Distributions

Consider two distributions $P(x|\theta)$ with an old parameter setting $\theta$, and $P(x|\widetilde{\theta})$ with a new parameter setting $\widetilde{\theta}$. Following Amari [Ama85] one may see the exponential family $\mathcal{F}_G$ as a manifold. The parameters $\theta$ and $\widetilde{\theta}$ represent two points on this manifold. Several measures of distance (divergence) between these two points have been proposed in the literature. Amari introduced $\alpha$ divergences [Ama85], and other related "distances" were introduced by Csiszar [Csi91] known as $f$ divergences (we use the letter $h$ below). Also Chernoff distances and Renyi's $\alpha$ information are related [Ama85]. These divergences all have the following general form:

$$E_\theta \, h\left(\frac{P_G(x|\widetilde{\theta})}{P_G(x|\theta)}\right),$$

where $h(.)$ is some continuous convex function. In the context of the exponential families the natural parameters $\widetilde{\theta}$ and $\theta$ are names for two distributions. "Good" measures of closeness or divergence between them should remain invariant when the two distributions are named by their expectation parameters $\widetilde{\mu}$ and $\mu$, respectively.

We limit our discussion to two choices of $h$. For $h(z) = -\ln z$ we get the relative entropy

$$\begin{aligned} E_\theta \ln \frac{P(x|\theta)}{P(x|\widetilde{\theta})} &= G(\widetilde{\theta}) - G(\theta) - (\widetilde{\theta} - \theta) \cdot \mu \\ &= F(\mu) - F(\widetilde{\mu}) - (\mu - \widetilde{\mu}) \cdot \widetilde{\theta}. \end{aligned}$$

The second equality follows from the duality (2.3). Choosing $h(z) = z \ln z$ gives the "opposite" relative entropy

$$\begin{aligned} E_{\widetilde{\theta}} \ln \frac{P(x|\widetilde{\theta})}{P(x|\theta)} &= G(\theta) - G(\widetilde{\theta}) - (\theta - \widetilde{\theta}) \cdot \widetilde{\mu} \\ &= F(\widetilde{\mu}) - F(\mu) - (\widetilde{\mu} - \mu) \cdot \theta. \end{aligned}$$

These two entropies are called $-1$ and $+1$ divergences by Amari [Ama85], respectively.

## 2.3 Bregman Divergence and its Properties

We just defined the notion of divergence between two exponential distributions using the cumulant function $G$ whose Hessian is positive definite. We now give a slightly more general definition of divergence due to Bregman [Bre67]: For any arbitrary differentiable strictly convex function $G$,

$$\Delta_G(\widetilde{\theta}, \theta) := G(\widetilde{\theta}) - G(\theta) - (\widetilde{\theta} - \theta) \cdot \mu,$$
where as before $\mu = \nabla_\theta G(\theta) = g(\theta)$  (2.6)

Note that $\Delta_G(\widetilde{\theta}, \theta)$ equals $G(\widetilde{\theta})$ minus the first two terms of the Taylor expansion of $G(\widetilde{\theta})$ around $\theta$. In other words, $\Delta_G(\widetilde{\theta}, \theta)$ is the tail of the Taylor expansion of $G(\widetilde{\theta})$ beyond the linear term.

Bregman divergences (or distances) have been used extensively in convex optimization. An important tool is a notion of projections w.r.t. the divergences. Bregman's essential method is to pick from a convex set of allowable parameters the one of minimal distance to the current parameter. In other words the current hypotheses is projected onto the set of allowable parameters. Some mild additional assumption on $G(\theta)$ assures the uniqueness of the projections. With this assumption a generalized Pythagorean Theorem can be proven for Bregman divergences [Bre67, CL81, Csi91, JB90, HW98]. The latter theorem contradicts the triangular inequality which is the reason why we use the term "divergence" instead of "distance".

In the context of learning these divergences were rediscovered in [KW97, HKW95, KW98]. Related divergences are used in [GLS97]. Projections have recently been applied in [HW98] for the case when the underlying model shifts over time and the projections w.r.t. the divergences are used to keep the parameters of the algorithm in reasonable convex regions. This aids the recovery process when the underlying model shifts.

### Properties of Bregman divergences

Using our notation $\mu = g(\theta) = \nabla_\theta G(\theta)$

1. $\Delta_G(\widetilde{\theta}, \theta) \geq 0$ and equality holds iff $\widetilde{\theta} = \theta$.
2. Often $\Delta_G(\widetilde{\theta}, \theta) \neq \Delta_G(\theta, \widetilde{\theta})$, when $\widetilde{\theta} \neq \theta$.
3. $\nabla_{\widetilde{\theta}} \Delta_G(\widetilde{\theta}, \theta) = g(\widetilde{\theta}) - g(\theta) = \widetilde{\mu} - \mu$.
4. $\Delta_G(\widetilde{\theta}, \theta)$ is strictly convex in the first argument and not necessarily convex in its second argument.
5. If $H(\theta) = G(\theta) + a \cdot \theta + b$, for $a \in \mathbf{R}^d$ and $b \in \mathbf{R}$, then $\Delta_H(\widetilde{\theta}, \theta) = \Delta_G(\widetilde{\theta}, \theta)$.

6. $\Delta_G(\widetilde{\theta}, \theta) = \Delta_F(\mu, \widetilde{\mu})$,
   where $F(\mu) := \theta \cdot \mu - G(\theta)$.

The first five properties are immediate and the last follows from the definition of the dual function $F$. Note that the order of the arguments in $\Delta_G(\widetilde{\theta}, \theta)$ and $\Delta_F(\mu, \widetilde{\mu})$ is switched as we move from $\theta$ to the dual parameter $\mu$ and that $\Delta_F(\mu, \widetilde{\mu})$ is again convex in its first argument.

## 3 Density Estimation with the Exponential Families

In this section we will analyze the performance of on-line algorithms for density estimation when the underlying distribution belongs to the exponential family with cumulant function $G(\theta)$. We start with the motivation of the update of the algorithms and then develop techniques for proving relative loss bounds. We make the most obvious choices. The loss of parameter $\theta$ on the example $x_t$ is

$$-\ln P(x_t|\theta) = G(\theta) - \theta \cdot x_t - \ln P_0(x_t).$$

We denote this loss as $L_t(\theta)$ and use $L_{1..t}(\theta) = \sum_{q=1}^{t} L_q(\theta)$ as shorthand for the loss in the first $t$ trials. We now motivate the algorithms with the (Bregman) divergence w.r.t. the cumulant function $G$.

The off-line algorithm has at hand all $T$ examples $\{x_1, x_2, \ldots, x_T\}$ and minimizes

$$\eta_B^{-1} \Delta_G(\theta, \theta_1) + L_{1..T}(\theta). \qquad (3.1)$$

Note that this maximizes the posterior when the prior is chosen to be the conjugate prior and $\eta_B^{-1} > 0$ is a measure of the importance of the prior. Alternatively one can think of $\theta_1$ as an initial parameter estimate based on some hypothetical examples seen before the first real example $x_1$ and $\eta_B^{-1} \geq 0$ as the number of those examples. The parameter $\eta_B$ measures the trade-off between staying close to the initial parameter $\theta_1$ and minimizing the loss on the $T$ real examples. In the example of Gaussian density estimation discussed in the introduction we have zero initial examples and thus $\eta_B^{-1} = 0$.

The gradient of (3.1) is

$$\eta_B^{-1}(\mu - \mu_1) + \sum_{t=1}^{T}(\mu - x_t)$$

Setting it to zero gives the best expectation parameter of the off-line algorithm:

$$\mu_B = (\eta_B^{-1} + T)^{-1}(\eta_B^{-1}\mu_1 + \sum_{t=1}^{T} x_t). \qquad (3.2)$$

The on-line algorithm starts with the same parameter setting $\mu_1 = g(\theta_1)$ and with a learning rate $\eta_1$. At the end of trial $t$ the examples $\{x_1, x_2, \ldots, x_t\}$ are available and the on-line algorithm minimizes:

$$V_{t+1}(\theta) = \eta_1^{-1}\Delta_G(\theta, \theta_1) + L_{1..t}(\theta), \qquad (3.3)$$

with $\eta_1^{-1} \geq 0$. Thus the expectation parameter of the on-line algorithm at end of trial $t$ is

$$\mu_{t+1} = (\eta_1^{-1} + t)^{-1}(\eta_1^{-1}\mu_1 + \sum_{q=1}^{t} x_q). \qquad (3.4)$$

We can also get a recursive version of the on-line update by minimizing

$$\eta_t^{-1}\Delta_G(\theta, \theta_t) + L_t(\theta) \qquad (3.5)$$

at the end of trial $t$. This gives

$$\mu_{t+1} = \mu_t - \eta_t(\mu_{t+1} - x_t), \qquad (3.6)$$

We always choose the learning rate $\eta_t$ at trial $t$ as

$$\eta_t = (\eta_1^{-1} + t - 1)^{-1}. \qquad (3.7)$$

With this choice the above update coincides with the expanded version (3.4). Thus this makes the update *permutation invariant*, i.e. independent of the order of the past instances. The following alternate recursive forms are useful:

$$\mu_{t+1} = \eta_{t+1}(\eta_t^{-1}\mu_t + x_t) \qquad (3.8)$$
$$\mu_{t+1} = \mu_t - \eta_{t+1}(\mu_t - x_t) \qquad (3.9)$$

The recursive forms (3.6) and (3.9) show that the on-line update may be seen as gradient descent with different learning rates. The update (3.6) uses the gradient of the loss at $\mu_{t+1}$, while (3.9) uses the gradient of the loss evaluated at $\mu_t$. Finally, (3.8) expresses the new expectation parameter $\mu_{t+1}$ as a convex combination of the current expectation parameter and example.

The hypothesis $\theta_t$ at trial $t$ represents a summary of the information gained from seeing the previous $t - 1$ examples. Minimizing (3.5) may be seen as striking a tradeoff between two different goals: one is to stay close to the current hypothesis and the other is to stay close to (or, equivalently, minimize the loss on) the last example seen. The first wants to preserve the learning gained from the past as it is encapsulated by the current hypothesis, while the second wants to learn from the last example in case it sees it again. The balance between the two goals is governed by the learning rate. Closeness to the current hypothesis is measured by the divergence $\Delta_G$. This way of motivating on-line updates was used extensively in [KW97]. Here we let the trade-off parameter $\eta_t$ change with time.

We distinguish between two initializations of $\eta_1$ for our on-line algorithm (given in (3.4) or (3.6-3.9)):

**Incremental off-line algorithm:**
$$\eta_1^{-1} = \eta_B^{-1} \text{ and } \eta_t = (\eta_B^{-1} + t - 1)^{-1}.$$

**Forward algorithm:**
$$\eta_1^{-1} = \eta_B^{-1} + 1 \text{ and } \eta_t = (\eta_B^{-1} + t)^{-1}.$$

Note that the hypothesis $\mu_{t+1}$ obtained by the incremental off-line algorithms at the end of trial $t$ coincides with the hypothesis of the batch algorithm using the initial $t$ examples. In the forward algorithm the inverse of the learning rate is shifted forward by one. This algorithm is inspired by the work of Vovk [Vov97] who uses a similar forward algorithm for linear regression.

Both algorithms start with the same initial expectation parameter $\mu_1$. However, in the second trial the incremental off-line algorithm predicts with $\mu_2 = (\eta_B^{-1} + 1)^{-1}(\eta_B^{-1}\mu_1 + x_1)$ while the forward algorithm predicts with $\mu_2 = (\eta_B^{-1} + 2)^{-1}((\eta_B^{-1} + 1)\mu_1 + x_1)$. Since the algorithms already have different predictions in the second trial their relative loss bounds are in general incomparable.

Recall that relative loss bounds are upper bounds on the difference between the total loss of the on-line algorithm and the total loss of the off-line algorithm, that is

$$\sum_{t=1}^{T} L_t(\theta_t) - \inf_{\theta} \left( \eta_B^{-1} \Delta_G(\theta, \theta_1) + L_{1..T}(\theta) \right). \quad (3.10)$$

We first develop concise forms for the two terms of the above difference.

**Lemma 3.1**
$$\begin{aligned}
\sum_{t=1}^{T} L_t(\theta_t) &= \sum_{t=1}^{T} \eta_{t+1}^{-1} \Delta_F(\mu_{t+1}, \mu_t) \\
&\quad + \eta_1^{-1} F(\mu_1) - \eta_{T+1}^{-1} F(\mu_{T+1}) \\
&\quad - \sum_{t=1}^{T} \ln P_0(x_t).
\end{aligned}$$

**Proof:**   We essentially rewrite the loss of the algorithm using the dual function $F$ and the update (3.9):

$$\begin{aligned}
& L_t(\theta_t) + \ln P_0(x_t) \\
&= G(\theta_t) - x_t \theta_t \\
&= \mu_t \theta_t - F(\mu_t) - x_t \theta_t \\
&= -\eta_{t+1}^{-1}(\mu_{t+1} - \mu_t)\theta_t - F(\mu_t) \\
&= \eta_{t+1}^{-1} \left( F(\mu_{t+1}) - F(\mu_t) - (\mu_{t+1} - \mu_t)\theta_t \right) \\
&\quad -\eta_{t+1}^{-1} F(\mu_{t+1}) + (\eta_{t+1}^{-1} - 1)F(\mu_t) \\
&= \eta_{t+1}^{-1} \Delta_F(\mu_{t+1}, \mu_t) - \eta_{t+1}^{-1} F(\mu_{t+1}) + \eta_t^{-1} F(\mu_t).
\end{aligned}$$

Finally we sum over $t$.   □

Recall that $\eta_B^{-1} \Delta_G(\theta, \theta_1) + L_{1..T}(\theta)$ is minimized at the natural parameter $\theta_B$ and expectation parameter $\mu_B$.

**Lemma 3.2**
$$\begin{aligned}
&\inf_{\theta} \left( \eta_B^{-1} \Delta_G(\theta, \theta_1) + L_{1..T}(\theta) \right) \\
&= \eta_B^{-1} F(\mu_1) - (\eta_B^{-1} + T)^{-1} F(\mu_B) - \sum_{t=1}^{T} \ln P_0(x_t).
\end{aligned}$$

**Proof:**   We rewrite the argument of the inf:

$$\begin{aligned}
& \eta_B^{-1} \Delta_G(\theta, \theta_1) + L_{1..T}(\theta) + \sum_{t=1}^{T} \ln P_0(x_t) \\
&= \eta_B^{-1} \Delta_F(\mu_1, \mu) + \sum_{t=1}^{T} (G(\theta) - x_t \cdot \theta)) \\
&= \eta_B^{-1} \{ F(\mu_1) - F(\mu) - (\mu_1 - \mu) \cdot \theta \} \\
&\quad + \sum_{t=1}^{T} (\mu \cdot \theta - F(\mu) - x_t \cdot \theta) \\
&= \eta_B^{-1} F(\mu_1) - (\eta_B^{-1} + T) F(\mu) \\
&\quad + (\eta_B^{-1} + T)\mu \cdot \theta - (\eta_B^{-1}\mu_1 + \sum_{t=1}^{T} x_t) \cdot \theta.
\end{aligned}$$

Setting $\mu = \mu_B$, which is the argmin of the inf (see (3.2)), the last two terms above cancel and the lemma follows.   □

The following lemma expresses the difference of the losses in one trial i.t.o. divergences.

**Lemma 3.3** For trial $t = 1, \ldots, T$ and $\theta \in \Theta$,

$$\begin{aligned}
L_t(\theta_t) - L_t(\theta) &= \eta_t^{-1} \Delta_G(\theta, \theta_t)) - \eta_{t+1}^{-1} \Delta_G(\theta, \theta_{t+1}) \\
&\quad + \eta_{t+1}^{-1} \Delta_G(\theta_t, \theta_{t+1})
\end{aligned}$$

**Proof:**   Since $V_{t+1}(\theta) - \eta_{t+1}^{-1} G(\theta)$ is linear (3.3), for $t = 0, \ldots, T$, we have by Property (5) that

$$\Delta_{V_{t+1}}(\widetilde{\theta}, \theta) = \Delta_{\eta_{t+1}^{-1} G}(\widetilde{\theta}, \theta) = \eta_{t+1}^{-1} \Delta_G(\widetilde{\theta}, \theta).$$

Thus the three divergences used in the lemma become:

$$\Delta_{V_t}(\theta, \theta_t) - \Delta_{V_{t+1}}(\theta, \theta_{t+1}) + \Delta_{V_{t+1}}(\theta_t, \theta_{t+1}).$$

We now simply expand the three divergences:

$$\begin{aligned}
& V_t(\theta) - V_t(\theta_t) - (\theta - \theta_t) \cdot \nabla V_t(\theta_t) \\
-\ & V_{t+1}(\theta) + V_{t+1}(\theta_{t+1}) + (\theta - \theta_{t+1}) \cdot \nabla V_{t+1}(\theta_{t+1}) \\
+\ & V_{t+1}(\theta_t) - V_{t+1}(\theta_{t+1}) - (\theta_t - \theta_{t+1}) \cdot \nabla V_{t+1}(\theta_{t+1}).
\end{aligned}$$

Since $\theta_t = \mathrm{argmin}_\theta V_t(\theta)$, for $t = 1,\ldots,T+1$, all gradients in the above are zero and the three divergences simplify to

$$V_t(\theta) - V_{t+1}(\theta) + V_{t+1}(\theta_t) - V_t(\theta_t).$$

Since $V_t(\theta) - V_{t+1}(\theta) = -L_t(\theta)$ and $V_{t+1}(\theta_t) - V_t(\theta_t) = L_t(\theta_t)$, the lemma follows. □

We now sum the equality of the lemma. Note that the second and third term on the r.h.s. telescope.

$$\sum_{t=1}^{T} L_t(\theta_t) - \left(L_{1..T}(\theta) + \eta_1^{-1}\Delta_G(\theta,\theta_1)\right)$$

$$= -\eta_{T+1}^{-1}\Delta_G(\theta,\theta_{T+1}) + \sum_{t=1}^{T} \eta_{t+1}^{-1}\Delta_G(\theta_t,\theta_{t+1})$$

By plugging $\theta = \theta_B$ into the above equality we get the following relative loss bound (This bound can also be proven by subtracting the expressions of the lemmas (3.1) and (3.2)).

**Theorem 3.4** *For both algorithms*

$$\sum_{t=1}^{T} L_t(\theta_t) - \left(L_{1..T}(\theta_B) + \eta_B^{-1}\Delta_G(\theta_B,\theta_1)\right)$$

$$= (\eta_1^{-1} - \eta_B^{-1})\Delta_G(\theta_B,\theta_1) - \eta_{T+1}^{-1}\Delta_G(\theta_B,\theta_{T+1})$$

$$+ \sum_{t=1}^{T} \eta_{t+1}^{-1}\Delta_G(\theta_t,\theta_{t+1})$$

Note that for the incremental off-line algorithm $\eta_1^{-1} = \eta_B^{-1}$ and $\theta_{T+1} = \theta_B$. Thus the first two terms on the r.h.s. of the above equality are zero.

### 3.1 Density Estimation with a Bernoulli

Now the examples are coin flips in $\{0,1\}$ and the cumulant function is $G(\theta) = \ln(1 + e^\theta)$. The parameter transformations are $\mu = g(\theta) = \frac{e^\theta}{1+e^\theta}$ and $\theta = f(\mu) = \ln\frac{\mu}{1-\mu}$ and the dual to $G(\theta)$ is $F(\mu) = \mu\ln\mu + (1-\mu)\ln(1-\mu)$. The loss (negative log-likelihood) i.t.o. the expectation parameter is $L_t(\theta) = -x_t\ln\mu - (1-x_t)\ln(1-\mu)$. It is easy to see that for the Bernoulli distribution the total loss of our on-line algorithms is permutation invariant (independant of the order of the instances). Consider the forward algorithm with $\eta_B^{-1} = 0$, $\eta_1^{-1} = 1$ and $\mu_1 = 1/2$. In this case the batch algorithm uses the fraction of ones $\mu_B = \frac{\sum_{t=1}^{T} x_t}{T}$ and the forward algorithm uses $\mu_{t+1} = \frac{1/2 + \sum_{q=1}^{t} x_q}{t+1}$. By a simple induction one can show that the total loss is

$$\sum_{t=1}^{T} L_t(\theta_t) = \ln T! - \ln \prod_{t=1}^{T\mu_B}(t-1/2) - \ln\prod_{t=1}^{T-T\mu_B}(t-1/2).$$

Thus

$$\sum_{t=1}^{T} L_t(\theta_t) - \sum_{t=1}^{T} L_t(\theta_B) = \ln \Gamma(T+1)$$
$$- \ln\Gamma(T\mu_B + 1/2) - \ln\Gamma(T - T\mu_B + 1/2) - \ln(\pi)$$
$$+ T(\mu_B \ln\mu_B + (1-\mu_B)\ln(1-\mu_B)).$$

This expression was first derived in [Fre96] using integration. With standard approximations of the $\Gamma$ function the above difference between the losses can be upper bounded [Fre96] by $\frac{1}{2}\ln(T+1) + 1$.

### 3.2 Density Estimation with a Gaussian

For a vector $x \in \mathbf{R}^d$, $x^2$ is shorthand for the real $x'x$. For Gaussian density estimation the cumulant function is $G(\theta) = \theta^2/2$, and the parameter transformation $g(\theta) = \mu$ is the identity function, i.e. $\theta = \mu$. The loss and the distance are quadratic, i.e. $L_t(\theta_t) = (\mu_t - x_t)^2/2 + c$ and $\Delta_G(\theta_t, \theta_{t+1}) = (\mu_{t+1} - \mu_t)^2/2$. Here $c$ is a constant and we choose $c = 0$. We use the on-line updates (e.g. (3.6), (3.9)) to rewrite the divergence between $\theta_t$ and $\theta_{t+1}$:

$$2\eta_{t+1}^{-1}\Delta_G(\theta_t,\theta_{t+1}) = \eta_{t+1}^{-1}(\mu_{t+1} - \mu_t)^2$$
$$\stackrel{(3.9)}{=} (\mu_{t+1} - \mu_t)(x_t - \mu_t)$$
$$= (\mu_{t+1} - \mu_t)(x_t + \mu_{t+1}) - (\mu_{t+1} - \mu_t)(\mu_{t+1} + \mu_t)$$
$$\stackrel{(3.6)}{=} \eta_t(x_t - \mu_{t+1})(x_t + \mu_{t+1}) + \mu_t^2 - \mu_{t+1}^2$$
$$= \eta_t(x_t^2 - \mu_{t+1}^2) + \mu_t^2 - \mu_{t+1}^2.$$

In this case the difference between the total losses of the on-line algorithm and the off-line algorithm (see Theorem 3.4) becomes:

$$(\eta_1^{-1} - \eta_B^{-1})(\mu_1 - \mu_B)^2/2 - \eta_{T+1}^{-1}(\mu_{T+1} - \mu_B)^2/2$$
$$+ \sum_{t=1}^{T} \eta_t(x_t^2 - \mu_{t+1}^2)/2 + \mu_1^2/2 - \mu_{T+1}^2/2$$
$$= \sum_{t=1}^{T} \eta_t x_t^2/2 - \sum_{t=1}^{T-1} \eta_t \mu_{t+1}^2/2$$
$$- (1 + \eta_T)\mu_{T+1}^2/2 + \mu_1^2/2$$
$$+ (\eta_1^{-1} - \eta_B^{-1})(\mu_1 - \mu_B)^2/2 - \eta_{T+1}^{-1}(\mu_{T+1} - \mu_B)^2/2.$$

We now assume $\mu_1 = 0$ for the rest of this section. For the incremental off-line algorithm (when $\eta_1 = \eta_B$ and $\mu_{T+1} = \mu_B$) the last three terms of the above are zero and the fourth last is negative. Also for the forward update (when $\eta_1^{-1} = \eta_B^{-1} + 1$ and $\mu_{T+1} = \frac{\eta_B^{-1}+T}{\eta_B^{-1}+T+1}\mu_B$), the last four terms can be shown to equal zero. Thus the difference of the total losses can be bounded by

$$\sum_{t=1}^{T} \eta_t x_t^2/2 - \sum_{t=1}^{T-1} \eta_t \mu_{t+1}^2/2. \qquad (3.11)$$

This bound holds for both algorithms. For the forward algorithm the difference is equal to the above.

It is easy find two examples for which the above difference depends on the order in which the examples are presented. By dropping the last sum in (3.11) we obtain an upper bound on the difference which is is now independent of the order (i.e. permutation invariant). The remaining sum can be crudely bounded by using

$$\sum_{t=1}^{T} \eta_t = \sum_{t=1}^{T} 1/(\eta_1^{-1} + t - 1) \leq \int_{\eta_1^{-1}-1}^{\eta_1^{-1}+T-1} (1/x)\,dx$$
$$= \ln((\eta_1^{-1} + T - 1)/(\eta_1^{-1} - 1))$$

and $X^2 = \max_{t=1}^{T} x_t^2$. Thus the difference of the losses for both algorithms is at most

$$\frac{X^2}{2} \ln(1 + \frac{T}{\eta_1^{-1} - 1}). \qquad (3.12)$$

### 3.3  Density Estimation with a Gamma

The Gamma density with shape parameter $\alpha$ and inverse scale parameter $\beta$ is

$$P(x|\alpha, \beta) = \frac{e^{-x\beta} x^{\alpha-1} \beta^\alpha}{\Gamma(\alpha)}, \quad \text{for } x, \alpha, \beta > 0.$$

This is a member of the exponential family with natural parameter $\theta = -\alpha/\beta$. So the density above i.t.o. $\theta$ can be written as

$$P(x|\theta, \alpha) = e^{\alpha(x\theta + \ln(-\theta))} \frac{\alpha^\alpha x^{\alpha-1}}{\Gamma(\alpha)}.$$

We assume $\alpha$ is known and fixed. Note that $\alpha^{-1}$ is called the dispersion parameter. The parameter $\alpha$ scales the loss and the divergences. So for the sake of simplicity we drop $\alpha$ just as we ignored the fixed variance in the case of Gaussian density estimation.

The cumulant function for the Gamma density is $G(\theta) = -\ln(-\theta)$ and the parameter transformations are $g(\theta) = -1/\theta = \mu$ and $f(\mu) = -1/\mu = \theta$. The dual convex function $F(\mu)$ becomes $-1 - \ln \mu$. In this section we only bound the divergence between $\theta_t$ and $\theta_{t+1}$. This essentially leads to a relative loss bound for the incremental off-line algorithm (See Theorem 3.4):

$$\eta_{t+1}^{-1} \Delta_G(\theta_t, \theta_{t+1}) = \eta_{t+1}^{-1} \Delta_F(\mu_{t+1}, \mu_t)$$
$$= \eta_{t+1}^{-1} (\ln \frac{\mu_t}{\mu_{t+1}} + \frac{\mu_{t+1}}{\mu_t} - 1).$$

We now use the update (3.9) and the notation $r_t = x_t/\mu_t > 0$ to rewrite the divergence:

$$\eta_{t+1}^{-1} \Delta_F(\mu_{t+1}, \mu_t)$$

$$= \eta_{t+1}^{-1} \left( \ln(\frac{\mu_t}{\mu_t - \eta_{t+1}(\mu_t - x_t)}) - \frac{\eta_{t+1}(\mu_t - x_t)}{\mu_t} \right)$$
$$= \eta_{t+1}^{-1} \ln(1 + \frac{1 - r_t}{\eta_t^{-1} + r_t}) + r_t - 1$$
$$\leq \eta_{t+1}^{-1} \frac{1 - r_t}{\eta_t^{-1} + r_t} + r_t - 1$$
$$\leq \frac{(1 - r_t)^2}{\eta_t^{-1} + r_t} \leq \eta_t (1 - r_t)^2$$

If $x_t \leq \mu_t$, then $(1 - r)^2 \leq 1$; if $x_t > \mu_t$, then $(1-r)^2 \leq (\frac{x_t}{\mu_t})^2$. Thus if $X = \max_{t=1}^T x_t$ and $Z = \min(\{x_t : 1 \leq t \leq T\} \cup \{\mu_1\})$, then $(1 - r)^2 \leq X^2/Z^2$ and the sum of the last terms can be bounded by

$$\frac{X^2}{Z^2} \sum_{t=1}^{T} \eta_t = \frac{X^2}{Z^2} O(\log T).$$

### 3.4  The General Case

We rewrite the divergence between $\theta_t$ and $\theta_{t+1}$:

$$\eta_{t+1}^{-1} \Delta_G(\theta_t, \theta_{t+1}) = \eta_{t+1}^{-1} \Delta_F(\mu_{t+1}, \mu_t)$$
$$= \eta_{t+1}^{-1}(F(\mu_{t+1}) - F(\mu_t) - (\mu_{t+1} - \mu_t) \cdot f(\mu_t)).$$

After doing a second order Taylor expansion of $F(\mu_{t+1})$ at $\mu_t$, this last term equals

$$\eta_{t+1}^{-1}(\mu_{t+1} - \mu_t)' F_{ij}(\widetilde{\mu}_t)(\mu_{t+1} - \mu_t)/2$$
$$= \eta_{t+1}(\mu_t - x_t)' F_{ij}(\widetilde{\mu}_t)(\mu_t - x_t)/2,$$

where $\widetilde{\mu}_t$ is a convex combination of $\mu_t$ and $\mu_{t+1}$. If $F_{ij}(\mu)$ is constant then we have a Gaussian. The general case may be seen as a local Gaussian with the time-varying curvature $F_{ij}(\mu)$. Any reasonable methods have to proceed on a case-by-case basis using the form of $F_{ij}(\mu)$ (which is the inverse of the variance function $V(\mu)$). Recall that $\eta_{t+1} = O(1/t)$. So summing the last term should give $\log(T)$-style bounds.

## 4  Linear Regression

In on-line regression the learning algorithms receive a sequence of *labeled* examples $(x_1, y_1), (x_2, y_2), \ldots, (x_T, y_T)$ in $\mathbf{R}^d \times \mathbf{R}$. In linear regression the loss is $L_t(\theta) = (x_t \cdot \theta - y_t)^2/2$. The incremental off-line algorithm predicts with $x_t \cdot \theta_t$ right after receiving the instance $x_t$ but before receiving the label $y_t$. It chooses its parameter vector in $\mathbf{R}^d$ by minimizing

$$(\theta - \theta_1)' \eta_B^{-1}(\theta - \theta_1)/2 + \sum_{q=1}^{t-1} L_q(\theta),$$

where $\eta_B^{-1}$ is now some symmetric strictly positive definite matrix. Thus when $\theta_1 = 0$

$$\theta_{t+1} = \eta_t \sum_{q=1}^{t} x_q y_q, \qquad (4.1)$$

where $\eta_t = (\eta_B^{-1} + \sum_{q=1}^{t} x_q x_q')^{-1}$.

This algorithm is also known as on-line linear least squares or ridge regression. Vovk's forward algorithm [Vov97] uses the same update (4.1) but now

$$\eta_t = (\eta_B^{-1} + \sum_{q=1}^{t+1} x_q x_q')^{-1}.$$

In the full paper we will show how linear regression can be viewed as a generalization of Gaussian density estimation. The proof of the relative loss bounds are similar except that the learning rates are now covariance matrices. In the corresponding learning rates for density estimation each product $x_q x_q'$ is one. When $\theta_1 = 0$ the difference between the total loss of both on-line algoritms can be shown to be at most

$$\sum_{t=1}^{T} y_t^2 x_t' \eta_t x_t / 2 - \sum_{t=1}^{T-1} (\theta_t \cdot x_t)^2 x_t' \eta_t x_t / 2.$$

In the case of the forward algorithm the difference is exactly the above (See (3.11) for the corresponding expression in Gaussian density estimation). After dropping the second sum, the resulting bound on the difference of the losses becomes permutation invariant. If $y_t \in [-Y, Y]$, $x_{t,i} \in [-X, X]$, and $\eta_B^{-1} = aI$ for some $a > 0$, then the remaining first sum can be roughly upper bounded by $\frac{dY^2}{2} \ln(1 + \frac{TX^2}{a})$ (See (3.12) for the corresponding expression).

## 5 Relationship to expanded updates Bayes

There is an alternate method pioneered by Vovk for proving relative loss bounds. A distribution is maintained on the set of models (or experts) $\Theta$. The initial work only considered the case when $\Theta$ is finite [Vov90, LW94]. However the method can also be applied when $\Theta$ is infinite [Fre96, Vov97, Yam98]. The distribution at the end of trial $t$ has the form

$$P_{t+1}(\theta) \sim P_1(\theta) \, e^{-\eta \sum_{q=1}^{t} L_q(\theta)},$$

where $P_1$ is a prior and $\eta$ a learning rate. When $\eta = 1$ and $L_t(x_t) = -\ln(P(x_t|\theta))$ then $P_{t+1}$ is the posterior after seeing the first $t$ examples. The following type of inequality is the main part of the method:

$$\begin{aligned} L_A(t) &\leq -\frac{1}{\eta} \ln \int_\theta P_t(\theta) e^{-\eta L_t(\theta)} \\ &= -\frac{1}{\eta} \ln \int_\theta P_1(\theta) e^{-\eta L_{1..t}(\theta)} \\ &\quad + \frac{1}{\eta} \ln \int_\theta P_1(\theta) e^{-\eta L_{1..t-1}(\theta)} \end{aligned}$$

Here $L_A(t)$ denotes the loss of the algorithm at trial $t$. The predictions of the algorithm are chosen so that the inequality holds for all sequences of examples. Also the larger the learning rate $\eta$, the better the resulting relative loss bounds. The same learning rate is used in all trials. The inequality is often tight when the examples lie on the boundary of the set of possible examples. By summing the inequality over all trials one gets the bound

$$\sum_{t=1}^{T} L_A(t) \leq -\frac{1}{\eta} \ln \int_\theta P_1(\theta) e^{-\eta L_{1..T}(\theta)} \qquad (5.1)$$

The integral is now bounded using Laplace's method of integration using the best off-line model as a saddle point [Fre96].

Freund applied Vovk's method to the Bernoulli distribution [Fre96] and obtained the same bound that is sketched in Section 3.1. For the Bernoulli the total loss of algorithms is permutation invariant. The examples are binary and thus always lie on the boundary and so the bounds are exact. For Gaussian density estimation the boundary of the example space is $\{x_t : x_t^2 = X^2\}$, for some positive $X$. Now the total loss of the algorithms is not permutation invariant.

Note that the r.h.s. of inequality (5.1) is permutation invariant. So any method that is based on inequality (5.1) can only lead to permutation invariant bounds. However Vovk's method does not seem to lead to the best possible permutation invariant bounds. Our method gives better permutation invariant bounds on the total loss of the forward algorithm after dropping some terms from an exact reformulation of the total loss. Only when all examples lie on the boundary then both methods agree. We also avoid the use of involved integration methods needed for (5.1) [Vov97, Fre96].

## 6 Conclusion

In this paper we presented techniques for proving relative loss bound for density estimation with the exponential family. We gave a number of examples of how to apply our methods including the case of linear regression. More examples will be given in the full paper. For an exponential density with cumulant function $G$ we use the divergence $\Delta_G(\widetilde{\theta}, \theta)$ as the measure of distances between the parameters $\widetilde{\theta}$ and $\theta$. Thus the loss $L_t(\theta) = -\ln P_G(x_t, \theta)$ and divergence are based on the same function $G$. However Lemma 3.3 and the methodology for proving relative loss bounds can be generalized to the case when the loss and the divergence use different strictly convex functions.

The parameters maintained by our algorithms are invariant to permuting the past examples. However the total on-line loss of the algorithms may not be permutation invariant. So an adversary could use this fact and present the examples in an order disadvantageous to the learning algorithm. We are currently exploring methods for making the total loss of the algorithms permutation invariant.

## Acknowledgements

Thanks to Leonid Gurvitz for introducing us to Bregman divergences and to Juergen Forster and Claudio Gentile for valuable comments. Juergen also found an alternate proof of the relative loss bounds for linear regression.